\documentclass[10pt,twocolumn,letterpaper]{article}

\usepackage{amsmath}
\usepackage[pagenumbers]{iccv} 
\usepackage{float}
\usepackage{algorithm}
\usepackage{mathtools}

\usepackage{algpseudocode}

\newcommand{\E}{\mathbb{E}}
\definecolor{iccvblue}{rgb}{0.21,0.49,0.74}
\usepackage[pagebackref,breaklinks,colorlinks,allcolors=iccvblue]{hyperref}

\def\paperID{9834} 
\def\confName{ICCV}
\def\confYear{2025}

\title{ROCM: RLHF on consistency models}

\author{Shivanshu Shekhar\\
University of Illinois Urbana-Champaign\\
{\tt\small shekhar6@illinois.edu}
\and
Tong Zhang\\
University of Illinois Urbana-Champaign\\
{\tt\small tozhang@illinois.edu}
}

\begin{document}
\maketitle
\begin{abstract}
Diffusion models have revolutionized generative modeling in continuous domains like image, audio, and video synthesis. However, their iterative sampling process leads to slow generation and inefficient training, challenges that are further exacerbated when incorporating Reinforcement Learning from Human Feedback (RLHF) due to sparse rewards and long time horizons. Consistency models address these issues by enabling single-step or efficient multi-step generation, significantly reducing computational costs.

In this work, we propose a direct reward optimization framework for applying RLHF to consistency models, incorporating distributional regularization to enhance training stability and prevent reward hacking. We investigate various $f$-divergences as regularization strategies, striking a balance between reward maximization and model consistency. Unlike policy gradient methods, our approach leverages first-order gradients, making it more efficient and less sensitive to hyperparameter tuning. Empirical results show that our method achieves competitive or superior performance compared to policy gradient based RLHF methods, across various automatic metrics and human evaluation. Additionally, our analysis demonstrates the impact of different regularization techniques in improving model generalization and preventing overfitting.

\end{abstract}    
\section{Introduction}
\label{sec:intro}

\begin{figure}[!ht]
  \centering
  \includegraphics[width=\linewidth]{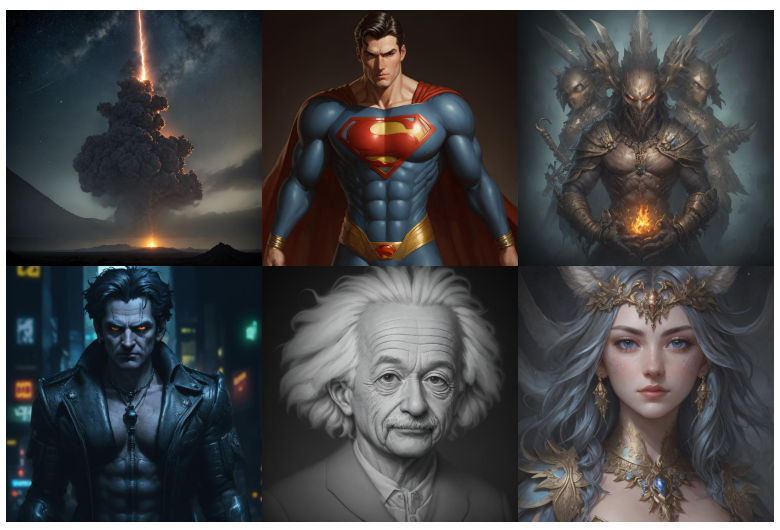}
  \caption{\textbf{Examples of images generated by the
model aligned using the KL divergence regularization constraint and HPS reward model}}
  \label{fig:intro}
\end{figure}

Diffusion models have brought about significant advancements in the modeling of continuous domains, including chemical molecule design \cite{mattergen}, audio generation \cite{audio}, text-to-image synthesis \cite{sd}, and video generation \cite{sora}. These models have demonstrated remarkable success across various applications, showcasing their versatility and potential. However, a notable challenge with diffusion models is their slow generation process. The iterative nature of the diffusion process means that each sample generation involves multiple steps, often making it difficult to train these models in an end-to-end manner. Researchers usually have to resort to approximations in the learning pipeline \cite{DDPO, DiffusionDPO} or endure extensive training times to fine-tune diffusion models effectively \cite{DRaFT}.

This challenge becomes even more pronounced when applying reinforcement learning (RL) pipelines to fine-tune diffusion models. In these scenarios, rewards are provided only at the final step of the generation process and as the time horizon increases the resulting sparse reward signals can significantly hinder the training performance. To tackle this problem, we turn our attention to the performance of Reinforcement Learning from Human Feedback (RLHF) when applied to consistency models \cite{Consistency}. Consistency models, in contrast to diffusion models, offer the advantage of  efficient generation with a small number of steps. In practice, they can produce competitive results within 4-8 steps, compared to the 20-50 steps typically required by diffusion models, thus addressing the issue of slow generation to a large extent.

A key observation in our study is that simply maximizing the reward in the RLHF pipeline often leads to overfitting and reward hacking, as the trained model diverges significantly from the original model. Reward hacking arises when the generation policy strays too far from the reference model, producing samples that are substantially different from those used to train the reward model. In such an out-of-distribution regime, a high reward does not necessarily indicate high-quality outputs. Over-optimizing for the reward can therefore lead to poor-quality images that receive artificially inflated scores.
To counter this, it is common to alter the objective to not only maximize the reward but also to minimize the divergence between the current model and the reference model distributions. The reference model is generally set to the base model at the beginning of training. By experimenting with different $f$-divergence measures, we find that this form of regularization helps stabilize the training process, preventing the model from degenerating into reward hacking and ensuring more robust performance across various metrics.

While prior research has applied methods such as Proximal Policy Optimization (PPO) and its variations to both diffusion and consistency models \cite{RLCM, DiffusionDPO, DDPO, DPOK, D3PO}, our work emphasizes that such complex training approaches may not always be necessary. We show that using the {\em reparameterization trick} \cite{kingma2014auto}, we can directly optimize the regularized RLHF objective by backpropagating through the entire generation trajectory.  Our experiments consistently demonstrate that direct optimization of the RLHF objective can outperform the use of PPO both in training stability and efficiency. Furthermore, a user study corroborates the effectiveness of our approach, underscoring its potential as a simpler yet robust alternative for training consistency models.
Our contributions in this work can be summarized as follows:
\begin{itemize}
    \item We formulate and analyze the role of distributional regularization in RLHF for fine-tuning consistency models, demonstrating its impact on training stability, efficiency and reward alignment. \\
    
    \item We reformulate the RLHF optimization problem as a direct optimization objective by leveraging the reparameterization trick, allowing efficient backpropagation through the generation trajectory. This reformulation transforms a zero-order optimization problem into a first-order one, significantly enhancing optimization efficiency. Empirically, our results demonstrate that this approach achieves performance on par with or superior to policy gradient based methods while requiring substantially less hyperparameter tuning and enabling faster training. \\
    
    \item We conduct a comprehensive empirical analysis of various $f$-divergence measures for regularization, highlighting their influence on training stability and model performance. 
\end{itemize}

\section{Related Works}
\label{sec:formatting}

\textbf{Diffusion Models:}
Diffusion models have emerged as a powerful class of generative models for tasks involving modeling of continuous data distributions. Inspired by non-equilibrium thermodynamics, these models learn to reverse a stochastic process that gradually adds noise to data, effectively learning the data distribution by reversing this process during generation \cite{Thermo, DDPM}. The iterative nature of diffusion models, where samples are generated through a sequence of denoising steps, allows them to produce high-quality outputs. However, this multi-step generation process is computationally intensive, leading to long inference times. To address this, recent work has focused on developing more efficient variants, such as DDIM \cite{DDIM}, which accelerates sampling by reducing the number of steps while maintaining output quality. Despite these advancements, the slow sampling speed of diffusion models remains a significant limitation, particularly when integrated with reinforcement learning frameworks for fine-tuning, as storing gradients for every timestep remains memory-intensive, even when only fine-tuning the LoRA layers \cite{LoRA}.

\textbf{Consistency Models:}
Consistency models present an alternative approach to generative modeling by enabling single-step or few-step generation \cite{Consistency}. These models are trained to maintain consistency in their outputs across multiple forward passes, facilitating much faster sampling compared to traditional diffusion models. The core idea is to train a network capable of directly mapping noise from any point in time to the target data distribution in a single step. For multi-step generation, noise is added at each step to the predicted target distribution sample, followed by the re-application of the denoising network, which allows these models to achieve competitive results in a limited number of steps. This approach drastically reduces the computational overhead during inference, making it particularly advantageous for scenarios requiring rapid generation. The ability of consistency models to produce high-quality samples in just a few steps also makes them well-suited for integration with reinforcement learning frameworks, where efficient feedback is crucial. For instance, RLCM \cite{RLCM} employs PPO to fine-tune a consistency model. While their work is closely related to ours, the key distinction lies in our use of direct reward optimization instead of PPO; moreover our objective focuses on optimizing a regularized version of the reward signal.

\textbf{Reinforcement Learning from Human Feedback:}
RLHF has gained traction in aligning generative models with human preferences, particularly in cases where explicit reward signals are sparse or difficult to define \cite{RLHF}. By leveraging human feedback, RLHF helps models generate outputs that better match human expectations. In generative modeling, RLHF fine-tunes models using reward signals derived from human judgments, improving output quality and relevance. However, integrating RLHF with diffusion models presents challenges due to slow sampling, memory constraints, and sparse reward signals, typically provided only at the end of the generation process. 
Various adaptations, such as PPO and its variants, have been explored to mitigate these issues, but they require complex training procedures with extensive hyperparameter tuning. End-to-end RLHF training methods, like DRaFT \cite{DRaFT}, employ techniques such as gradient checkpointing and truncated backpropagation to manage computational overhead. Meanwhile, approaches based on Direct Preference Optimization (DPO) \cite{D3PO, DiffusionDPO, SPO} reformulate the training objective to decouple it from the generation steps, allowing optimization without storing per-step gradients. RLHF has also been utilized to enhance generation diversity \cite{SEEDPO}, though this is not the focus of our work.

\begin{figure*}[!htbp]
  \centering
  \includegraphics[width=0.85\textwidth]{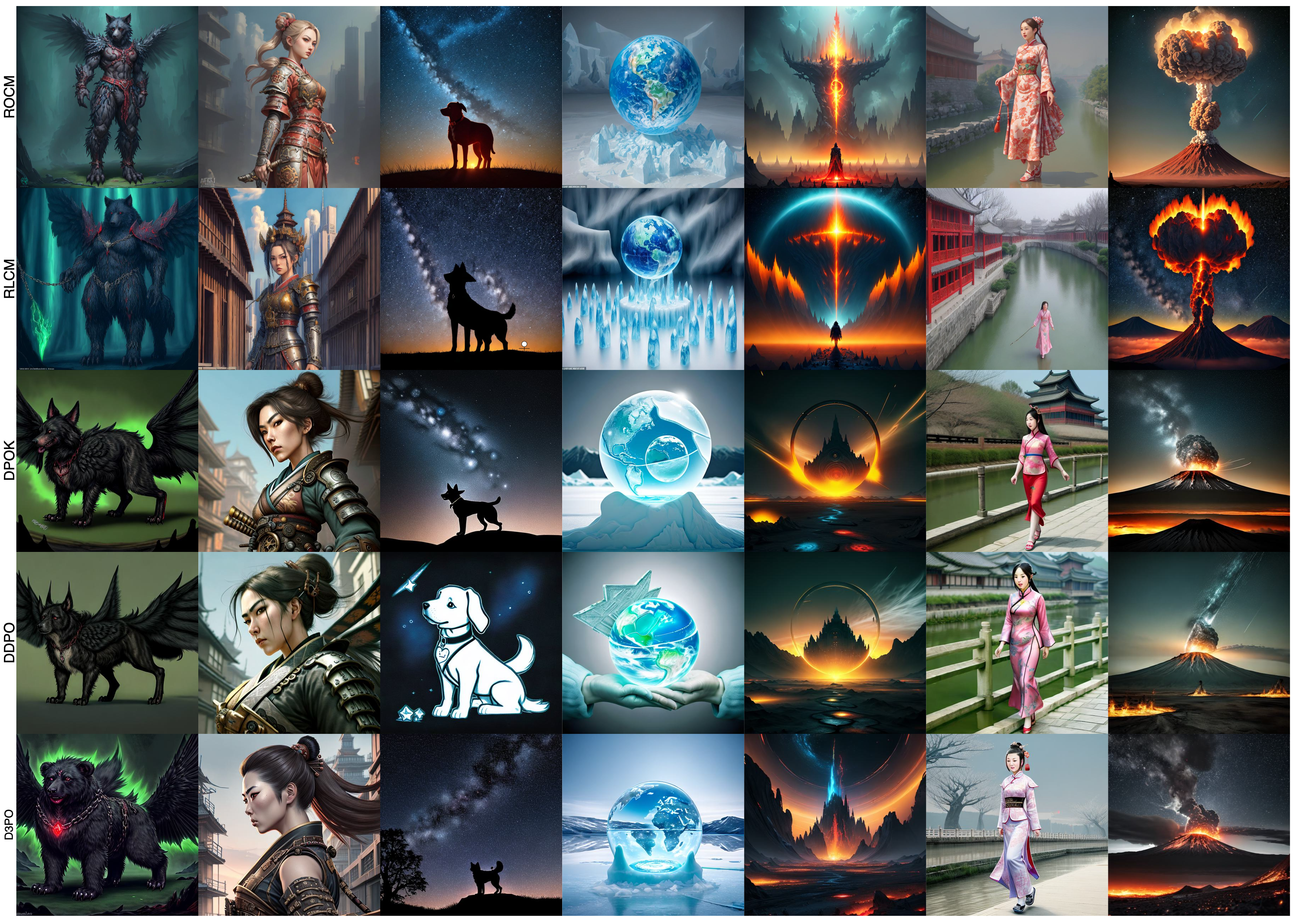}
  \caption{\textbf{Sample Images generated by our baselines and ROCM trained on HPSv2 as reward model.}}
  \label{fig:compare_main}
\end{figure*}
\section{Preliminaries \& Methodology}

\textbf{Consistency Models:}   
Diffusion models define a family of probability distributions \( p_t(x) \), parameterized by time \( t \in [0, T] \), where a clean data sample \( x_0 \sim p_0(x) \) undergoes a gradual noising process. At the terminal timestep \( T \), the data distribution converges to an isotropic Gaussian prior, i.e., \( x_T \sim \mathcal{N}(0, I) \). The forward diffusion process follows the transition kernel:
\begin{equation}
\label{eq:tr_ker}
    q_{t}(x_t|x_0) = \mathcal{N}(x_t;\alpha_tx_0, \sigma_tI),
\end{equation}

where \( \alpha_t, \sigma_t \) govern the noise schedule at each step. This stochastic process can equivalently be described by the following Stochastic Differential Equation (SDE) \cite{PFODE, dpm++, karras}:
\begin{equation}
    dx_t = f(t)x_t + g(t)dw_t,
\end{equation}

where \( w_t \) denotes standard Brownian motion, and \( f(t) \) and \( g(t) \) are functions of \( \alpha_t \) and \( \sigma_t \). The marginal distribution \( q_t(x) \) evolving under this forward-time SDE satisfies a corresponding reverse-time SDE, which can alternatively be formulated as an Ordinary Differential Equation (ODE) \cite{PFODE}:
\begin{equation}
    \label{eq:PFODE}
    \frac{d x_t}{d t} = f(t) x_t + \frac{1}{2} g^2(t)  \epsilon_{\theta}(x_t, t).
\end{equation}

Here, \( x_T \sim \mathcal{N}(0, I) \), and \( \epsilon_{\theta} \) represents a neural network trained to approximate the score function of \( q_t(x_t) \). This noise-prediction model can be enhanced using classifier-free guidance \cite{cfg}, modifying the predicted noise as:
\begin{equation}
    \label{eq:cfg}
    \hat{\epsilon}_{\theta}(x_t, \omega, c, t) = (1 + \omega)\epsilon_{\theta}(x_t, c, t) - \omega \epsilon_{\theta}(x_t, \phi, t),
\end{equation}

where \( c \) represents the conditioning input (typically a text prompt) and \( \omega \) is the guidance scale, which modulates the trade-off between sample diversity and specificity. Substituting \eqref{eq:cfg} into \eqref{eq:PFODE}, we obtain the Augmented Probability Flow ODE (APFODE) \cite{lcm}:
\begin{equation}
    \label{eq:APFODE}
    \frac{d x_t}{d t} = f(t) x_t + \frac{1}{2} g^2(t)  \hat{\epsilon}_{\theta}(x_t, \omega, c, t).
\end{equation}

Consistency models aim to accelerate generative sampling by learning a direct mapping from noisy samples to high-quality outputs in a single or few-step inference process. Unlike conventional diffusion models, which iteratively refine samples by solving \eqref{eq:PFODE}, consistency models approximate the solution trajectory of the ODE directly. Specifically, given two time steps \( t' > t \), if \( x_{t'} \sim p_{t'}(x) \), then by integrating \eqref{eq:PFODE}, one can obtain \( x_t \sim p_t(x) \) and ultimately recover \( x_0 \sim p_0(x) \). The consistency model, parameterized by \( f_{\theta} \), learns the mapping:
\begin{equation}
    f_{\theta}(x_t, t) = f_{\theta}(x_{t'}, t') = x_0,
\end{equation}
where $x_0$ is the solution of \eqref{eq:PFODE} at time $0$ starting from $x_t$ at time $t$.
ensuring that the recovered sample remains consistent across different time steps. Training is performed by minimizing a distance function \( d(f_{\theta}(x_t, t), f_{\theta}(x_{t'}, t')) \), often using the \( L_2 \) norm \cite{Consistency}. This encourages the generated samples to remain close to the true data distribution.

Following prior work \cite{Consistency}, we parameterize \( f_{\theta} \) as:
\begin{equation}
    f_{\theta}(x_t, t) = c_{\text{skip}}(t) x_t + c_{\text{out}}(t) F_{\theta}(x_t, t),
\end{equation}

where \( c_{\text{skip}}(t) \) and \( c_{\text{out}}(t) \) are differentiable functions with constraints \( c_{\text{skip}}(0) = 1 \) and \( c_{\text{out}}(0) = 0 \). The term \( F_{\theta}(x_t, t) \) represents a neural network with learnable parameters \( \theta \). Following \cite{lcm}, the classifier-free guidance scale \( \omega \) \cite{cfg} can be incorporated into the consistency function as $f_\theta(x_{t_k}, \omega, c, t_k)$ allowing the model to directly predict the solution to the APFODE \eqref{eq:APFODE}.

One of the primary advantages of consistency models is their ability to generate high-quality samples with significantly fewer inference steps. The probability flow trajectory is discretized into a sequence of \( K \) decreasing timesteps, \( T = t_K > t_{K-1} > \dots > t_1 = 0 \), where each step refines the sample towards \( x_0 \). This process defines a generation policy \( \pi_\theta \) that maps noisy inputs to high-fidelity outputs efficiently.

Using Eq. \ref{eq:tr_ker} we can write for any sample at time \( t_n \), $x_{t_n} = \alpha_{t_n} x_0 + \sigma_{t_n} z, \quad z \sim \mathcal{N}(0, I)$. Given a sample \( x_k \) at time \( t_k \), we approximate \( x_0 \) using a consistency function \( \tilde{x}_k = f_{\theta}(x_k, \omega, c, t_k) \). The next-step sample is then obtained as:
\begin{equation*}
    x_{k-1} = \alpha_{t_{k-1}} \tilde{x}_k + \sigma_{t_{k-1}} \epsilon_{k-1}, \quad \epsilon_{k-1} \sim \mathcal{N}(0, I).
\end{equation*}

This iterative procedure balances computational efficiency with output quality. Following \cite{lcm}, we integrate classifier-free guidance \cite{cfg} into the generation process, as summarized in Algorithm \ref{alg:cmg}. Notably, at the final step, \( x_0 \approx \tilde{x}_1 = f_{\theta}(x_1, \omega, c, t_1) \), since the parameters satisfy \( \alpha_0 = 1 \) and \( \sigma_0 = 0 \).

\textbf{Regularized RLHF:}  
RLHF aims to align the output of generative models with human preferences by using human-provided feedback as a reward signal.  In regularized RLHF, the objective function is augmented with a regularization term to ensure stable training and prevent overfitting or reward hacking. The regularized RLHF objective can be expressed as:
\begin{equation}
\mathcal{L}_{\text{RLHF}} = \mathbb{E}_{\tau \sim \pi_\theta} \left[ R(\tau) \right] + \beta \, \mathcal{D}(\pi_\theta \| \pi_{\theta_{\text{ref}}}) .
\label{eq:rlhf}
\end{equation}
Here \( \tau \) represents a trajectory sampled from the consistency model generation policy \( \pi_{\theta} \) according to Algorithm \ref{alg:cmg}.
\( R(\tau) \) denotes the reward associated with the trajectory which is usually given at the last step of generation, \( \mathcal{D}(\cdot \| \cdot) \) is a divergence measure between the current policy \( \pi_\theta \) and a reference policy \( \pi_{\theta_{\text{ref}}} \) that
corresponds to the pretrained model with parameter $\theta_{\text{ref}}$,
and \( \beta \) is a regularization coefficient.
Note that in this work, we assume that the reward model $R(\cdot)$ is already given, and our goal is to learn a suitable policy $\pi_\theta$ associated with the consistency model. 

The regularization term plays a critical role in stabilizing the training process. In large language model applications, it is known that without this term, the model may quickly overfit to the specific reward model, leading to suboptimal generalization \cite{ziegler2019fine,stiennon2020learning}. This paper shows that the same holds true for consistency models. Common choices for the divergence \( \mathcal{D} \) include Kullback-Leibler (KL) divergence, which measures the relative entropy between two distributions, Jensen-Shannon divergence, a symmetric version of KL divergence, and Hellinger squared distance, which provides a notion of distance between distributions and Fisher divergence which measures the distance between distributions by comparing their score functions. Each divergence has unique properties that can affect the training dynamics, and the choice of \( \mathcal{D} \) depends on the specific requirements of the task. Since our trajectory relies on multiple intermediate steps, which is analogous to the chain of thought steps in large language models, where conditional KL regularization is applied to each of the steps, 
we also aggregate divergence of conditional distributions $p(x_{k-1}|x_k,c)$ over these intermediate steps for $k=2,\ldots,K$. According to Algorithm \ref{alg:cmg}, this conditional distribution,
which we denote by $p_{k}(\cdot|\theta, x_k,c)$ is a Gaussian distribution:
\begin{equation}
p_k(\cdot|\theta,x_k,c) = \mathcal{N}(\alpha_{t_{k-1}} f_{\theta_{\text{ref}}}(x_k,\omega,c,t_k), \sigma_{t_{k-1}}^2) .
\label{eq:conditional}
\end{equation}
We have the following expression, with $\tau=\{(x_k, \tilde{x}_k)\}_{k=1}^K$: 
\begin{equation*}
\mathcal{D}(\pi_\theta \| \pi_{\theta_{\text{ref}}})
= \E_{\tau \sim \pi_\theta} \sum_{k=2}^K 
\mathcal{D}_f( p_k(\cdot|\theta,x_k,c)|| p_k(\cdot|\theta_{\text{ref}},x_k,c))
\end{equation*}
where $\mathcal{D}_f$ is a properly chosen $f$-diverge, which we will elaborate later. 
One important property of this distributional regularization function is that it can be reparameterized. 

In the standard RL method, the derivative of the expectation \( \mathbb{E}_{\tau \sim \pi_\theta} \left[ R(\tau) \right] \) with respect to $\theta$ is calculated using policy gradient or its variation such as PPO. In this work, we use a reparameterization approach, which is known to reduce variance compared to policy gradient, and is widely used in various prior work such as variational autoencoder  \cite{kingma2014auto}. To see that reparameterization can be applied, we know that the randomness of Algorithm~\ref{alg:cmg} only comes from $K$ Gaussian variables $\epsilon_K,\ldots,\epsilon_1$, which we may aggregate to a bigger Gaussian variable $\epsilon$. The trajectory $\tau$ from Algorithm~\ref{alg:cmg} can be considered as a function of $\epsilon$, $\theta$, and condition $c$:
$\tau = G(\theta,\epsilon,c)=\{x_k\}_{k=0}^K$, where we also take $x_k=G_k(\theta,\epsilon,c)$ for $k=0,\ldots,K$.
Using this notation, we can rewrite \eqref{eq:rlhf} in the following reparameterized form:
\begin{align}
\mathcal{L}_{\text{RLHF}}& = \mathbb{E}_{\epsilon \sim \mathcal{N}(0,I)} \Bigg[ R(G_0(\theta,\epsilon,c),c)  + \beta \, \sum_{k=2}^K \label{eq:rlhf-reparam}\\
&   D_f(p_k(\cdot|\theta,G_k(\theta,\epsilon,c),c)||
p_k(\cdot|\theta_{\text{ref}},G_k(\theta,\epsilon,c),c)\Bigg] \notag
\end{align}
We call this reformulation the direct optimization formulation for RLHF. The gradient with respect to $\theta$ can be calculated using backpropagation through the generation steps in Algorithm~\ref{alg:cmg}. Since the distribution involves a sequence of conditional Gaussians, the distributional regularization considered in \eqref{eq:rlhf-reparam} can be either computed in closed form, or can be estimated using another Gaussian reparameterization (see Table~\ref{tab:fD}). 
In this formulation, the reward function \( R(\tau) \) captures human preferences by assigning higher values to trajectories that align better with human feedback. 
It is a known reward function that depends on the final image $x_0$ and condition $c$, which we assume is differentiable with respect to $x_0$. 
The regularization coefficient \( \beta\) balances the influence of the reward and the regularization term, controlling the degree of adherence to the reference policy. In our experimentation we found that a good balance of regularization and reward is achieved by scaling the divergence to be one order of magnitude smaller than the rewards.

\textbf{\textit{f}-Divergence:} The \( f \)-divergence is a general class of divergence measures used to quantify the difference between two probability distributions. Given a convex function \( f(x): \mathbb{R}^+ \rightarrow \mathbb{R} \) that satisfies \( f(1) = 0 \), and two discrete distributions \( p_1 \) and \( p_2 \) defined over a common space \( \mathcal{X} \), the \( f \)-divergence between them is formulated as follows \cite{FD}:
\begin{equation}
    \mathcal{D}_f (p_1||p_2) = \mathbb{E}_{x\sim p_2} \bigg[f\bigg(\frac{p_1(x)}{p_2(x)}\bigg) \Bigg]    . \label{eq:f-div}
\end{equation}
%

The \( f \)-divergence framework generalizes several widely used divergence measures by selecting different functions \( f(x) \). Notable examples include the Kullback-Leibler (KL) divergence (both forward and reverse forms), Jensen-Shannon (JS) divergence, Fisher divergence and Hellinger distance, each of which serves a specific purpose in machine learning and probabilistic modeling.
Since our distributional regularization is only concerned with Gaussian distributions defined in \eqref{eq:conditional}, some of the standard $f$-divergence can be computed using the closed form solutions given in Table~\ref{tab:fD}. For JS-Divergence which has no closed form solution, we can use reparameterization again in \eqref{eq:f-div} to estimate it with the function 
$f(p_k(x_{k-1}|\theta_{\text{ref}},x_k,c)/p_k(x_{k-1}|\theta,x_k,c))$ at step $k$.
The resulting algorithm is given in Algorithm~\ref{alg:train}.

\begin{table*}[!htbp]
    \centering
    \renewcommand{\arraystretch}{1} 
    \resizebox{0.75\textwidth}{!}{%
    \begin{tabular}{cc}
        \hspace{-3cm}
        \begin{subtable}[t]{0.48\textwidth}
            \centering
            \begin{tabular}{c|c|c|c|c|c}
                \hline
                Method & PickScore & Aesthetic & CLIP & BLIP & ImageReward \\ 
                \hline
                No-Regularization & 21.288 & \textbf{6.538} & 0.247 & 0.463 & 0.298 \\
                JS-Divergence & 21.669 & 6.310  & 0.265 & \textbf{0.489} & 0.498 \\
                KL-Divergence & 21.690 & 6.394 & 0.263 & 0.479 & \textbf{0.651} \\
                Hellinger & \textbf{21.701} & 6.352 & 0.266 & 0.486 & 0.610 \\
                Fisher-Divergence & 21.598 & 6.366 & \textbf{0.267} & 0.481 & 0.540 \\
                RLCM & 21.393 & 6.067 & 0.265 & 0.482 & 0.505 \\
                DDPO & 21.179 & 6.011 & 0.263 & 0.484 & 0.504  \\ 
                D3PO & 21.451 & 5.975 & 0.266 & 0.482 & 0.550  \\ 
                DPOK & 21.285 & 5.996 & 0.265 & 0.488 & 0.518  \\
                \hline
            \end{tabular}
            \caption{Trained using HPSv2 as reward model \cite{HPS}}
        \end{subtable} &
        \hspace{3cm}
        \begin{subtable}[t]{0.48\textwidth}
            \centering
            \begin{tabular}{c|c|c|c|c|c}
                \hline
                Method & HPSv2 & Aesthetic & CLIP & BLIP & ImageReward \\ 
                \hline
                No-Regularization & 2.73 & 5.957 & 0.254 & 0.474 & 0.163 \\
                JS-Divergence & \textbf{2.95} & 6.307 & 0.268 & 0.483 & \textbf{0.513} \\
                KL-Divergence & 2.79 & 6.301 & 0.268 & 0.483 & 0.485 \\
                Hellinger & 2.86 & 6.297 & \textbf{0.269} & \textbf{0.485} & 0.472 \\
                Fisher-Divergence & 2.82 & \textbf{6.331} & 0.267 & 0.483 & 0.431 \\
                RLCM & 2.68 & 6.318 & 0.265 & 0.481 & 0.341 \\
                DDPO & 2.80 & 5.818 & 0.261 & 0.478 & 0.246 \\
                D3PO & 2.50 & 5.876 & 0.267 & 0.476 & 0.205  \\
                DPOK & 2.81 & 5.933 & 0.265 & 0.482 & 0.354 \\ 
                \hline  
            \end{tabular}
            \caption{Trained using PickScore as reward model \cite{PickScore}}
        \end{subtable} \vspace{0.5cm} \\ 
        
        \hspace{-3cm}
        \begin{subtable}[t]{0.48\textwidth}
            \centering
            \begin{tabular}{c|c|c|c|c|c}
                \hline
                Method & PickScore & Aesthetic & HPSv2 & BLIP & ImageReward \\ 
                \hline
                No-Regularization & 21.010 & 6.115 & 2.48 & 0.467 & 0.039 \\
                JS-Divergence & 21.157 & 6.296 & 2.73 & 0.468 & 0.225 \\
                KL-Divergence & 21.420 & 6.226 & 2.79 & 0.479 & 0.334 \\
                Hellinger & \textbf{21.516} & 6.308 & 2.81 & 0.481 & \textbf{0.444} \\
                Fisher-Divergence & 21.455 & 6.244 & 2.79 & 0.481 & 0.359 \\
                RLCM & 21.396 & \textbf{6.313} & 2.71 & 0.471 & 0.337 \\
                DDPO & 21.173 & 5.952 & 2.78 & 0.482 & 0.383\\
                D3PO &  21.362 & 5.944 & 2.86 & \textbf{0.486} & 0.424  \\
                DPOK & 21.243 & 5.965 & \textbf{2.88} & 0.483 & 0.401  \\
                \hline
            \end{tabular}
            \caption{Trained using CLIPScore as reward model \cite{ClipScore}}
        \end{subtable} &
        
        \hspace{3cm}
        \begin{subtable}[t]{0.48\textwidth}
            \centering
            \begin{tabular}{c|c|c|c|c|c}
                \hline
                Method & PickScore & HPSv2 & CLIP & BLIP & ImageReward \\ 
                \hline
                No-Regularization & 21.237 & 2.64 & 0.260 & 0.470 & 0.186 \\
                JS-Divergence & 21.547 & 2.82 & 0.267 & \textbf{0.485} & 0.478 \\
                KL-Divergence & 21.626 & 2.86 & 0.266 & 0.481 & 0.521 \\
                Hellinger & \textbf{21.713} & \textbf{2.89} & 0.267 & 0.484 & \textbf{0.531} \\
                Fisher-Divergence & 21.573 & 2.87 & 0.266 & 0.482 & 0.488 \\
                RLCM & 21.362 & 2.64 & 0.265 & 0.476 & 0.319 \\
                DDPO & 21.435 & 2.85 & 0.265 & 0.479 & 0.456 \\
                D3PO & 21.110 & 2.67 & 0.262 & 0.469 & 0.248 \\
                DPOK & 21.468 & 2.85 & \textbf{0.268} & 0.482 & 0.483\\ 
                \hline
            \end{tabular}
            \caption{Trained using Aesthetic Score as reward model \cite{aes_score}}
        \end{subtable}
    \end{tabular}
    }
    \caption{\textbf{Comparison of different regularization techniques across multiple reward models. Each model is trained separately using PickScore \cite{PickScore}, HPSv2 \cite{HPS}, CLIPScore \cite{ClipScore}, and Aesthetic Score \cite{aes_score}. Models are not evaluated on the reward function they were trained on to avoid bias. We also include additional evaluations using BLIPScore \cite{blip} and ImageReward \cite{imagereward}. The reported scores are computed on the \texttt{validation\_unique} split of the Pick-A-Pic V1 dataset \cite{PickScore}.}}
    \label{tab:results}
\end{table*}

\begin{figure*}[!htbp]
  \centering
  \includegraphics[width=0.85\textwidth]{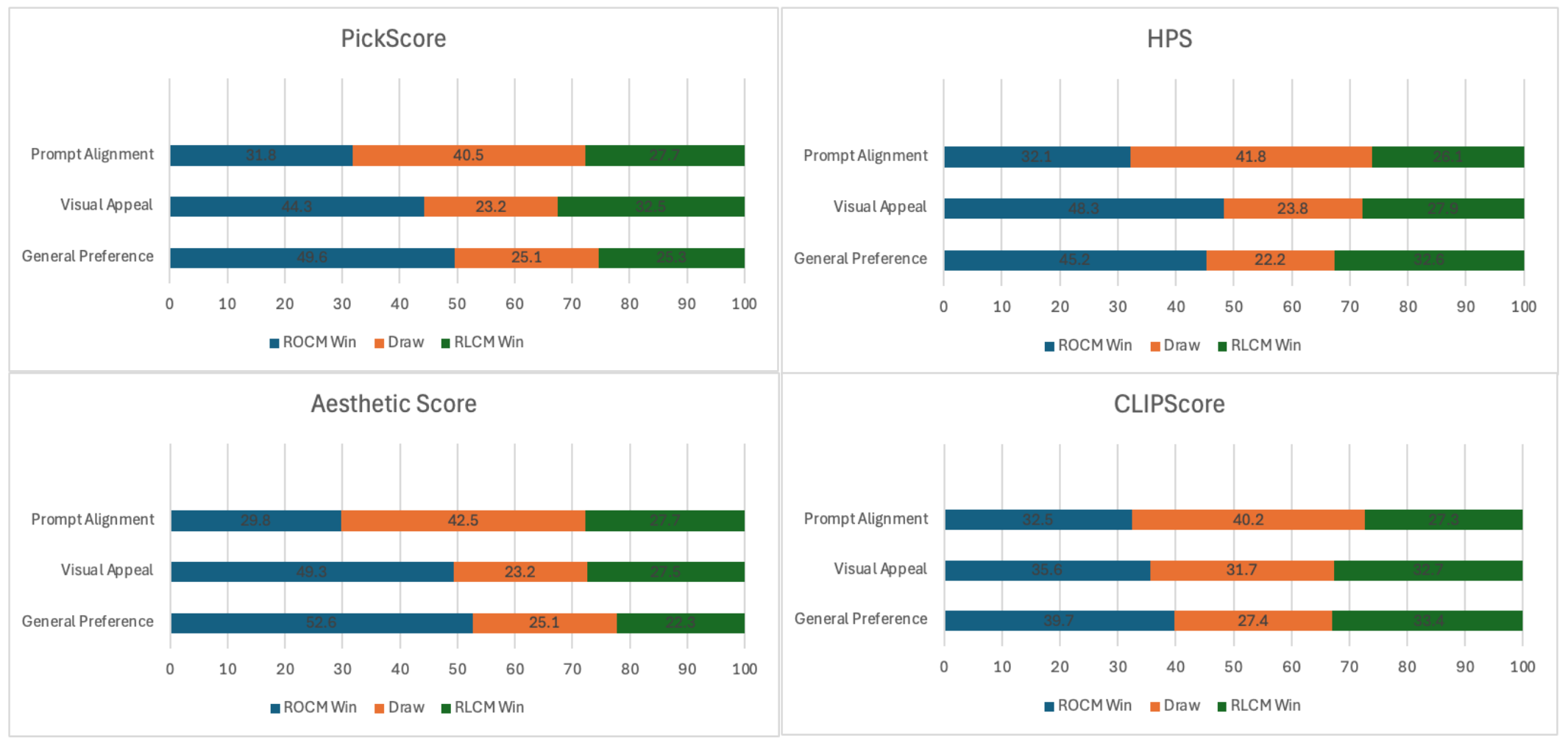}
  \caption{\textbf{User study comparing Our best models for each reward model with RLCM \cite{RLCM} fine-tuned on that reward model, we follow SPO \cite{SPO} and choose in total 300 randomly sampled prompts from Partiprompts \cite{Parti} and HPS \cite{HPS} we sample in the ratio of 1:2 respectively.}}
  \label{fig:2-Human Evaluation}
\end{figure*}

\begin{figure}[!htbp]
  \centering
  \includegraphics[width=0.85\linewidth]{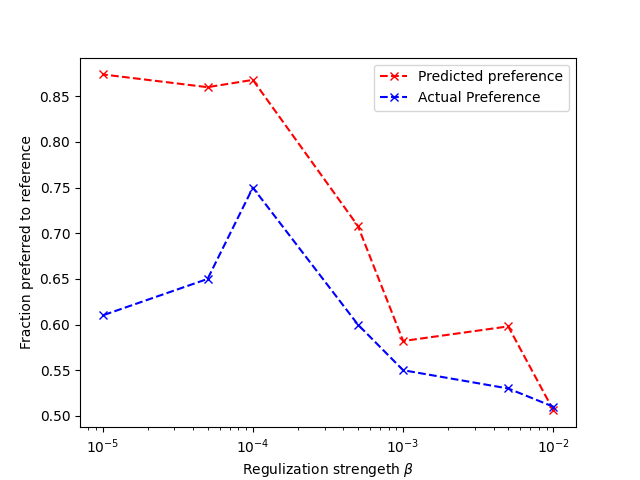}
  \caption{\textbf{As \(\beta\) decreases, we observe an initial improvement in model performance. However, with further reduction in \(\beta\), the actual preference reaches a peak and then begins to decline, indicating reward hacking.}}
  \label{fig:sens}
\end{figure}

\begin{figure*}[!htbp]
  \centering
  \includegraphics[width=0.88\textwidth]{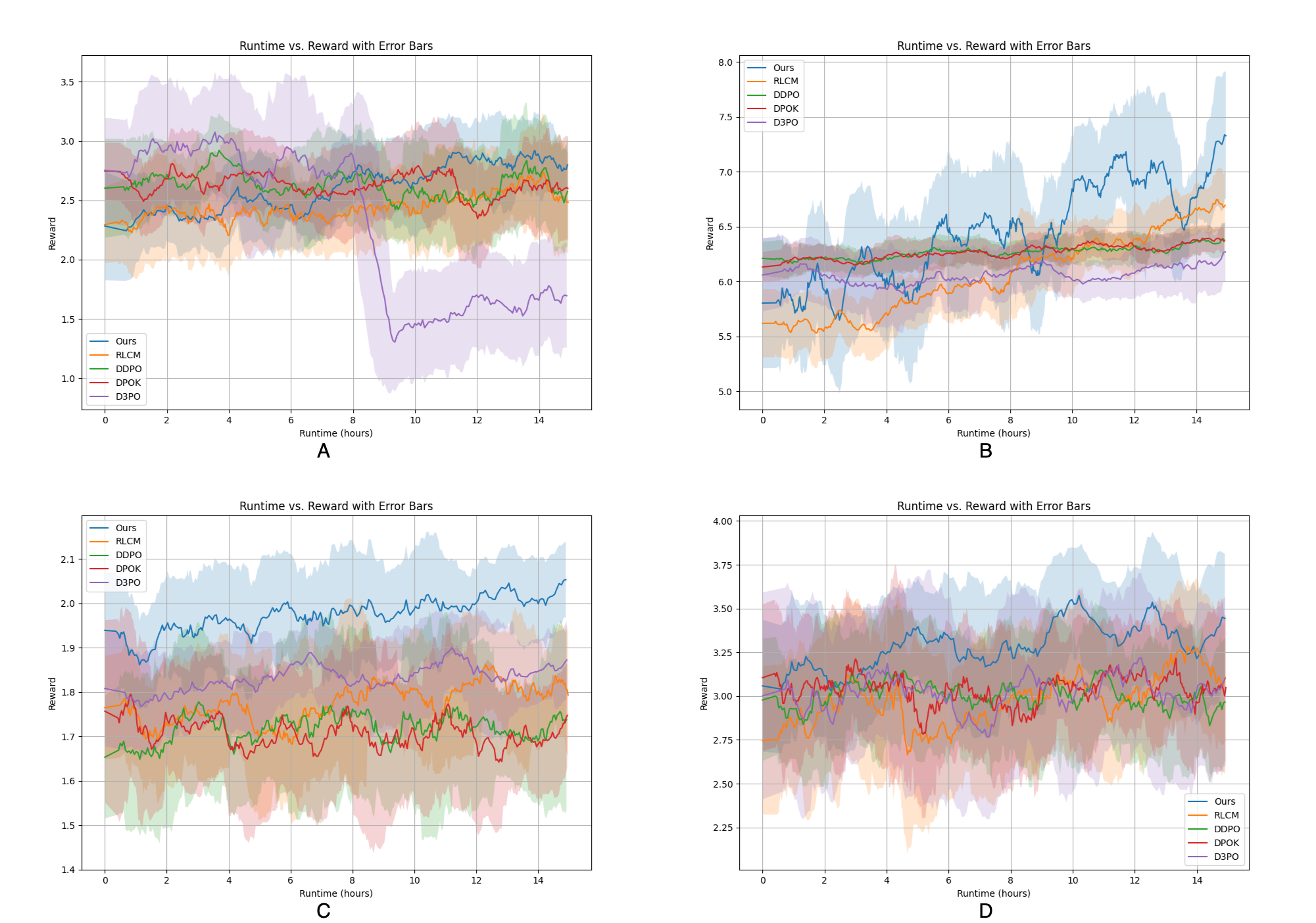}
  \caption{\textbf{This figure illustrates the training efficiency of each method, with Figures A, B, C, and D representing models trained using CLIPScore, Aesthetic Score, PickScore, and HPSv2, respectively. Our method consistently outperforms others in terms of training efficiency across different reward models. Notably, improvements are relatively minor for PickScore and CLIPScore. The limited gain in CLIPScore is expected, as it primarily aids in prompt alignment, while PickScore's lower sensitivity to image quality results in a smaller increase. In contrast, HPSv2 and Aesthetic Score exhibit significant improvements within just 15 GPU hours. We used a running average of window size 20 to arrive at the error bars and mean.}}
  \label{fig:Graphs}
\end{figure*}

\section{Experiments}
\label{experiments}
\textbf{Datasets:} To train our models in an online fashion—where each model is trained exclusively on its own generated data while being updated iteratively—we utilized 4,000 text prompts (without images) randomly sampled from the Pick-a-Pic V1 dataset, as employed in \cite{SPO}. This prompt dataset was used to fine-tune models with PickScore \cite{PickScore}, HPSv2 \cite{HPS}, CLIPScore \cite{ClipScore}, and Aesthetic Score \cite{aes_score}. Furthermore, for generating images in Fig: \ref{fig:compare}, we trained models using Aesthetic Score on a smaller set of 45 animal-related prompts, as in \cite{DDPO}.

For quantitative evaluation, we report results on 500 validation prompts present in the \texttt{validation\_unique} split of the Pick-a-Pic V1 dataset \cite{PickScore}, which was also utilized in \cite{SPO}. We train five models following Algorithm \ref{alg:train}, each incorporating a different regularization method: No regularization, KL-Divergence, JS-Divergence, Hellinger Distance, and Fisher Divergence. Our models are compared against baseline methods, including RLCM \cite{RLCM}, DDPO \cite{DDPO}, DPOK \cite{DPOK}, and D3PO \cite{D3PO}. Specifically, RLCM applies PPO to consistency models, DDPO employs PPO for diffusion models, DPOK utilizes policy gradient with a KL-regularized reward, and D3PO extends DPO \cite{DPO} to diffusion models.

\textbf{Implementation Details:} In our experiments, we employ an 8-step consistency model and a 20-step diffusion model, both utilizing classifier-free guidance \cite{cfg} with a guidance scale of $\omega=7.5$. For diffusion-based and consistency-based methods, we use Dreamshaper v7, a fine-tuned version of Stable Diffusion v1.5, along with its corresponding consistency model counterpart \cite{lcm} as base models respectively. These are further fine-tuned with trainable LoRA \cite{LoRA} layers. Specifically, we set the LoRA rank to 16 and $\alpha$ to 32 for consistency-based methods, as we observed that complex prompts from the Pick-a-Pic V1 dataset required a higher parameter capacity for better representation. For diffusion-based methods, we conducted a hyperparameter search to optimize performance. To ensure a fair comparison between all methods, we kept the learning rate the same for both models. Further experimental details can be found in the appendix (\ref{sec:details}).

We explore multiple $f$-divergences, namely KL-Divergence, Reverse-KL Divergence, Hellinger Squared Distance, and Jensen-Shannon Divergence, incorporating a hyperparameter $\beta$ to regulate regularization strength. The optimal values for these hyperparameters are detailed in the appendix (\ref{sec:details}). For all divergences except Jensen-Shannon, we utilize the closed-form solutions provided in Table \ref{tab:fD}. Since Jensen-Shannon Divergence lacks a closed-form solution, we resort to sampling for its computation.

\textbf{Evaluation Metrics:} For evaluation, we use 6 automated metrics: PickScore \cite{PickScore}, CLIPScore \cite{ClipScore}, HPSv2 \cite{HPS}, Aesthetic Score \cite{aes_score}, BLIPScore \cite{blip}, and ImageReward \cite{imagereward}. All metrics, except Aesthetic Score, are prompt-aware, while Aesthetic Score is prompt-agnostic and evaluates only the aesthetic quality using a linear estimator on a CLIP vision encoder. Each reward model has been trained on human preference data to approximate human image quality judgments.  PickScore and HPSv2 utilize a CLIP-based model trained on human preferences related to aesthetic quality and prompt-to-image alignment while ImageReward uses a BLIP-based model for the same. CLIPScore and BLIPScore focuses on prompt-to-image alignment only. 
Beyond automated metrics, we conduct a user study similar to \cite{SPO}. We recruited 10 participants to evaluate 300 image pairs generated by RLCM and our best-performing models for each reward model. The prompts for this study are randomly sampled from a mixture of the PartiPrompts dataset \cite{Parti} and the HPSv2 dataset \cite{HPS}, maintaining a 1:2 ratio.

\subsection{Results}  
We evaluate all regularization methods across different reward models, including RLCM \cite{RLCM}, D3PO \cite{D3PO}, DDPO \cite{DDPO}, and DPOK \cite{DPOK}, using the previously described evaluation metrics. The results are summarized in Table \ref{tab:results}, with each table representing models trained on a specific reward model. We exclude scores for the metric optimized in each table to avoid reporting inflated values due to potential overfitting. Additionally, we present training time vs. performance graphs in Fig: \ref{fig:Graphs}.

Across all tables and metrics, regularized-ROCM consistently outperform or match the performance of other approaches in automatic evaluations. In some cases, even the non-regularized-ROCM performs comparably or better than the baselines. Additionally, both regularized-ROCM and RLCM achieve higher scores on most metrics than their diffusion-based counterparts, highlighting the advantages of consistency models over diffusion models. This performance gap can be attributed to the challenges of fine-tuning diffusion models, which struggle with long diffusion trajectories and the sparse rewards encountered in RLHF. Fig: \ref{fig:Graphs} further illustrates that regularized-ROCM achieves superior scores in a shorter training duration compared to RLCM. This is likely due to the fact that PPO relies on noisy and unstable zeroth-order gradient approximations, leading to slower training, whereas our approach leverages more stable first-order gradients, enabling faster convergence and improved performance. Notably, both our methods and RLCM demonstrate superior training efficiency compared to their diffusion-based alternatives. Furthermore, as shown in Table \ref{tab:base_results}, diffusion-based methods exhibit a decline in performance across several automatic metrics when compared to their base model. Our experiments suggest that these models primarily optimize for the reward model they are trained on, often at the expense of performance on other metrics—a pattern also observed in RLCM. We observe the same phenomena in the non-regularized variant of ROCM, albeit to a lesser extent than baseline models. In contrast, regularized-ROCM models do not suffer from this issue, this suggests that both first-order gradients and regularization are essential for achieving superior overall performance.
The user study presented in Fig: \ref{fig:2-Human Evaluation} further validates our approach, showing that our methods significantly outperform RLCM across all reward models in terms of \textit{Visual Appeal} and \textit{General Preference}. While these improvements are substantial, the gains in \textit{Prompt Alignment} are relatively modest. Specifically, for the Aesthetic Score reward model, the prompt alignment remains nearly identical to that of RLCM. This is expected, as Aesthetic Score is a prompt-agnostic metric, meaning it does not inherently improve prompt alignment. In contrast, other reward models consider the prompt in their evaluation, leading to enhanced prompt alignment in those cases. We see a similar behavior for CLIPScore and \textit{Visual Appeal}, as CLIPScore is only meant to reward prompt-image alignment and not quality.

\subsection{Further Analysis}
\textbf{Effect of Regularization Strength ($\beta$):} Fig: \ref{fig:sens} illustrates the impact of the regularization strength parameter $\beta$ on model performance. We report results for each model after 10k iterations and use KL-Divergence for regularization. At higher values of $\beta$, the actual human preferences and the reward model (RM) predictions are nearly the same. As $\beta$ decreases, the RM's predicted preference increases more than the actual human preference. At $\beta = 10^{-4}$, human preference peaks and then declines, while the RM prediction continues to rise. This indicates overfitting to the reward model, which we refer to as reward hacking. Models trained with such overfitting generate artificially inflated scores, though the actual outputs remain noisy. Similar observations have been reported in \cite{LLMOVERFIT} for large language models.

\textbf{Effectiveness of reward models:} From Table \ref{tab:results} and Fig: \ref{fig:Graphs}, we observe that both PickScore and HPSv2 lead to substantial improvements in generation quality and prompt alignment. Notably, PickScore demonstrates lower sensitivity to image quality. Models trained with CLIPScore show limited improvements in image quality but offer notable benefits in prompt alignment, as expected, since it is designed for prompt-to-image alignment rather than image quality assessment. On the other hand, models trained with Aesthetic Score show significant improvements in image quality, but only a small improvement on prompt alignment. Since Aesthetic Score is prompt-agnostic, it can lead to models generating repetitive images that do not align with the given prompt.

\section{Conclusions \& Limitations}
In this paper, we demonstrated that Direct Reward Propagation for fine-tuning consistency models outperforms complex methods like PPO, which require extensive hyperparameter tuning. By utilizing the reparameterization trick, we optimized the regularized RLHF objective directly through backpropagation across the entire generation trajectory, improving training efficiency and stability. We explored the impact of distributional regularization in RLHF and showed that penalizing significant deviations from the initial model enhances both training stability and reward alignment. Our empirical results indicate that our approach not only surpasses prior methods in reward alignment and sample efficiency but also benefits from distributional regularization, which mitigates reward hacking effects that often occur when relying solely on reward scores as training signals.

Furthermore, we conducted a comparative analysis of different divergence measures for regularization, highlighting that while each affects the generated samples differently, all contribute to better generalization and resilience to overfitting compared to unregularized training. A limitation of our approach is that it requires differentiable reward signals, as it is first-order and relies on gradient-based optimization. Therefore, it is not directly applicable to tasks involving non-differentiable rewards, such as compressibility or incompressibility, where policy-gradient methods are still necessary. \par
{
    \small
    \bibliographystyle{ieeenat_fullname}
    \bibliography{main}
}

\clearpage
\setcounter{page}{1}
\maketitlesupplementary

\section{Additional Details}
\label{sec:details}
In this section, we provide the essential training details for our model. We trained our ROCM models using a batch size of 1 on 2 A6000 GPUs, with a gradient accumulation of 1, resulting in an effective batch size of 1. The learning rate was set to $6 \times 10^{-5}$. To ensure a fair comparison, we applied the same batch size and learning rate settings to our baseline models. For training and inference, we used 8 steps for consistency model-based methods and 20 steps for diffusion-based methods. The optimal $\beta$ values for training our models with each reward model are listed in Table \ref{tab:beta}. As base models, we used \texttt{SimianLuo/LCM\_Dreamshaper\_v7} for consistency models and \texttt{Lykon/dreamshaper-7} for diffusion models from huggingface.
 
\begin{table}[!htbp]
    \centering
    \begin{tabular}{|c|c|} 
        \hline
        Regularization & $\beta$ \\ 
        \hline
        JS-Divergence   & $2000$   \\  
        KL-Divergence   & $10^{-4}$   \\  
        Hellinger   & $0.5$   \\  
        Fisher-Divergence   & $5\times10^{-5}$ \\ 
        \hline
    \end{tabular}
    \caption{\textbf{Optimal $\beta$ values for ROCM}}
    \label{tab:beta}
\end{table}

\begin{figure*}[!htbp]
  \centering
  \includegraphics[width=\textwidth]{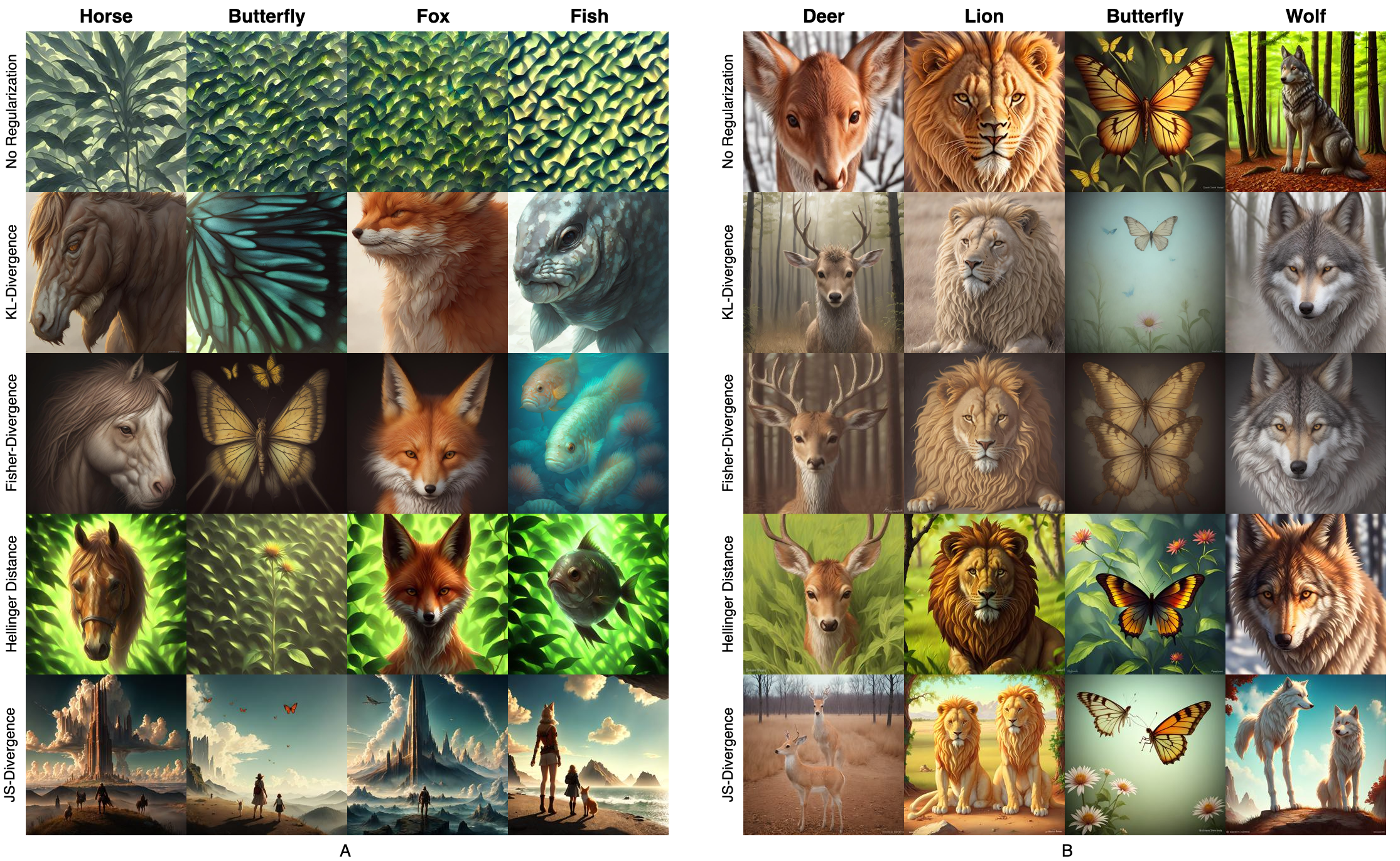}
  \caption{\textbf{Fig: A and B were generated using Aesthetic Score \cite{aes_score} as the reward model. Fig: B showcases the best results across all methods, highlighting how each divergence induces a unique style of image generation, with the reward model effectively accommodating these distinct styles. Fig: A illustrates the reward-hacked regions for each method, revealing that the overfitted images differ based on the divergence used. Notably, regularization proves beneficial, as the absence of regularization leads to erroneous, incoherent images, whereas regularized methods still produce legible images, albeit with suboptimal backgrounds.}}
  \label{fig:compare}
\end{figure*}

We observed another intriguing effect of regularization, as illustrated in Fig: \ref{fig:compare}. Each regularization method guides the model toward a specific generation style, while the Aesthetic Score reward model consistently assigns high scores to all of them, demonstrating its generality. Interestingly, we also find that unregularized methods produce highly noisy outputs, whereas regularized methods generate relatively coherent images, each exhibiting its own distinct style of overfitting.

\section{Algorithms}
\label{sec:rationale}
In Algorithm \ref{alg:cmg} we present the consistency multi-inference algorithm \cite{lcm} and in Algorithm \ref{alg:train} we present our training algorithm:

\begin{algorithm}[!htbp]
    \caption{Consistency Model K-step Generation}
    \label{alg:cmg}
    \begin{algorithmic}[1]
        \State Draw $x_{K} = \epsilon_K \sim \mathcal{N}(0, I)$
        \State $c\sim C$, where $C$ is the set of conditions (eg. prompts)
        \For{$k = K, \ldots, 1$}
            \State $\tilde{x}_k = f_\theta(x_{k}, \omega,c, t_k)$
            \State $\epsilon_{k-1} \sim \mathcal{N}(0, I)$
            \State $x_{k-1} = \alpha_{t_{k-1}} \tilde{x}_{k} + \beta_{t_{k-1}} \epsilon_{k-1}$
        \EndFor \\
        \Return $x_0$
    \end{algorithmic}
\end{algorithm}

\begin{algorithm}
\caption{Optimization with Divergence Regularization}
\label{alg:train}
\begin{algorithmic}[1]
\State Initialize parameters $\theta$
\State Set reference parameters $\theta_{\text{ref}} \coloneqq \theta$
\State Set batch size $B$ and regularization weight $\lambda$
\State Set reward model $R(\cdot)$ and divergence $\mathcal{D}$
\Repeat
    \State Sample a batch of conditions $\{c_i\}$ of size $B$
    \For{$i = 1, \ldots, B$}
        \State Sample noise $\epsilon^{(i)} \sim \mathcal{N}(0,I)$
       \State Generate trajectory $\{x^{(i)}_k\}_{k=0}^K= G(\theta, \epsilon^{(i)},c_i)$
        \State $R_i = R(x^{(i)}_0,c_i)$  
        \State $\mathcal{D}_i = \sum_{k=2}^K \mathcal{D}_f(p_k(\cdot|\theta,x_k^{(i)},c_i)||p_k(\cdot|\theta_{\text{ref}},x_k^{(i)},c_i))$
    \EndFor
    \State Update parameters using the objective:
        \[
        \theta \leftarrow \theta + \eta \nabla_\theta \left( \frac{1}{B} \sum_{i=1}^B \Big[R_i - \lambda \mathcal{D}_i \Big] \right)
        \]
\Until{Convergence}
\end{algorithmic}
\end{algorithm}

\section{$f$-divergence}
In Table: \ref{tab:fD} we present the table of $f$-divergences with their actual formula and closed form solution when the distributions are assumed to be Gaussian with different means and same standard deviation.
\begin{table*}[!htbp]
    \centering
    \small
    \setlength{\tabcolsep}{5 pt}
    \begin{tabular}{|c|c|c|}
    \hline
    $f$-divergence & $f(x)$ & $D(\mathcal{N}(\mu_1, \sigma^2 I), \mathcal{N}(\mu_2,\sigma^2 I))$ \\
    \hline\hline
    KL-Divergence & $x\log{x}$ & $\frac{\|\mu_1 - \mu_2 \|_2^2}{2\sigma^2}$\\
    Hellinger & $(\sqrt{x} - 1)^2$ & $1 - \exp{\frac{\|\mu_1 - \mu_2 \|_2^2}{8\sigma^2}}$\\
    JS-Divergence & $\frac{1}{2}(x\log{\frac{2x}{x+1}} + \log(\frac{2}{x+1})$ & N/A \\
    Fisher Divergence & $\left\|  \log x \right\|^2 dx$ & $\frac{\|\mu_1 - \mu_2 \|_2^2}{\sigma^4}$\\
    \hline
    \end{tabular}
    \caption{\textbf{This table summarizes the commonly used $f$-divergence. Here $x = \frac{p(t)}{q(t)}$. JS-Divergence doesn't have a closed form solution for two Gaussian distributions}}
    \label{tab:fD}
\end{table*}

\begin{table*}[!htbp]
    \centering
    \small
    \setlength{\tabcolsep}{5 pt}
    \begin{tabular}{|c|c|c|c|c|c|c|}
    \hline
    Method & PickScore & Aesthetic & HPSv2 & CLIP & BLIP & ImageReward \\
    \hline\hline
    Diffusion & 21.590 & 6.246 & 0.282 & 0.266 & 0.477 & 0.331 \\
    LCM & 21.181 & 6.005 & 0.261 & 0.254 & 0.456 & 0.023 \\
    \hline
    \end{tabular}
    \caption{\textbf{Performance metrics of baseline models for diffusion and consistency based methods.}}
    \label{tab:base_results}
\end{table*}

\end{document}